\documentclass{elsarticle}

\usepackage{microtype}
\usepackage{graphicx}
\usepackage{subfigure}
\usepackage{booktabs}
\usepackage{hyperref}
\usepackage{multirow}
\usepackage{amsmath}
\usepackage{amssymb}
\usepackage{mathtools}
\usepackage{amsthm}
\usepackage[capitalize,noabbrev]{cleveref}

\theoremstyle{plain}
\newtheorem{theorem}{Theorem}[section]
\newtheorem{proposition}[theorem]{Proposition}

\theoremstyle{definition}

\theoremstyle{remark}

\newcommand{\ie}{\textit{i}.\textit{e}.}
\newcommand{\etc}{\textit{etc}.}

\journal{Journal of \LaTeX\ Templates}

\begin{document}

\begin{frontmatter}

\title{Neural Collapse Inspired Attraction-Repulsion-Balanced Loss for Imbalanced Learning}

\author[zju]{Liang Xie}
\author[jd]{Yibo Yang}
\author[zju]{Deng Cai}
\author[zju,fabu]{Xiaofei He}

\address[zju]{State Key Lab of CAD \& CG, Zhejiang University, Hangzhou, China}
\address[jd]{JD Explore Academy, Beijing, China}
\address[fabu]{Fabu Inc., Hangzhou, China}

\begin{abstract}
Class imbalance distribution widely exists in real-world engineering. However, the mainstream optimization algorithms that seek to minimize error will trap the deep learning model in sub-optimums when facing extreme class imbalance. It seriously harms the classification precision, especially in the minor classes. The essential reason is that the gradients of the classifier weights are imbalanced among the components from different classes. In this paper, we propose Attraction-Repulsion-Balanced Loss (ARB-Loss) to balance the different components of the gradients. We perform experiments on large-scale classification and segmentation datasets, and our ARB-Loss can achieve state-of-the-art performance via only one-stage training instead of 2-stage learning like nowadays SOTA works.
\end{abstract}


\end{frontmatter}


\section{Introduction}

Class imbalance exists in almost all real-world scenes, and we call it long-tailed distribution when the imbalance is very extreme. Due to that standard gradient descent optimization algorithm seeking the solution to make the loss as low as possible, the learned algorithm tends to perform better on the major classes while ignoring the learning on those minor classes. It will trap the model in the sub-optimum. 

\begin{figure}[ht]
	\vskip 0.2in
	\begin{center}
		\centerline{\includegraphics[width=0.8\columnwidth]{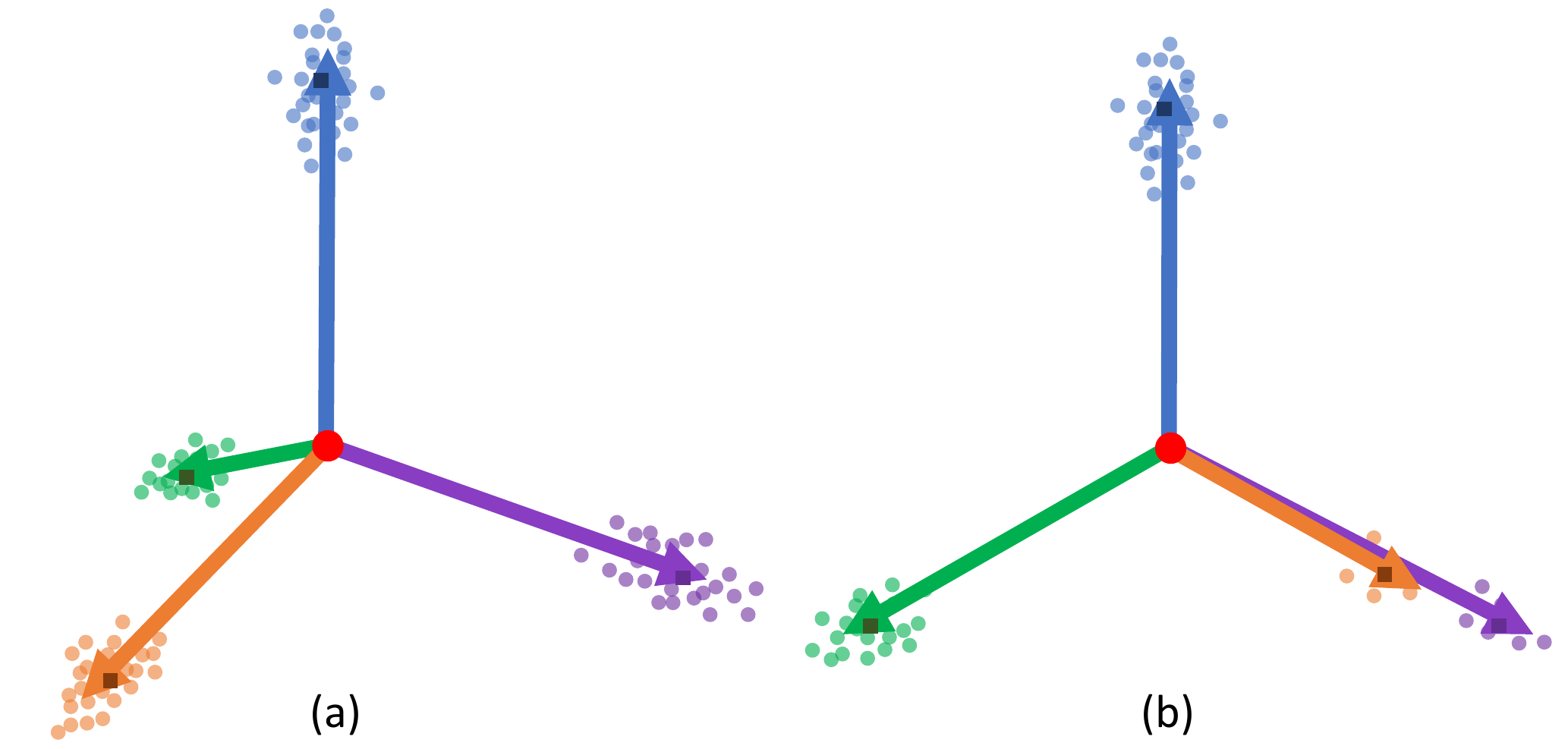}}
		\caption{(a) \textbf{\textit{Neural Collapse}}: when trained on the balanced data, the classifier weights and features are dually collapsed into an ETF geometric structure (Note that the figure depicts a 3D ETF structure, where the arrows are equal-length like the molecular structure of methane.); (b) \textbf{\textit{Minority Collapse}}: when the cardinalities of some classes (the \textit{orange} and \textit{purple} classes) are much lower than others (the \textit{green} and \textit{blue} classes), the weights corresponding to these rare classes collapsed into similar direction. (The arrow represents the classifier weights, the light-color spot represents the head of the feature vector the dark-color square represents the feature mean; each color represents a class.)}
		\label{fig:neural_minority_collapse}
	\end{center}
	\vskip -0.2in
\end{figure}

Many works have recently emerged to mitigate the performance drop under the long-tailed distribution. Some works \cite{chawla2002smote,buda2018systematic} re-sample the data to mitigate the class imbalance. However, over-sampling might cause the model to overfit these classes, while under-sampling might bring the risks of removing the important samples. Another line of works \cite{huang2016learning,rota2017loss,khan2017cost,cui2019class,cao2019learning,tan2020equalization} designs specific loss function to weight different classes according to the class cardinality. However, for the large-scale dataset, the re-weighting cost functions make deep models difficult to optimize. Additionally, these methods rely on hyper-parameters, which must be adjusted according to the specific data. Recently, many methods use two-stage training to decouple the representation learning, and the class-balanced training of the classifier \cite{cao2019learning,kang2019decoupling,zhou2020bbn,zhong2021improving}. They first use instance-balanced sampling to learn discriminative features, and then they fix the backbone network and train the classifier via class-balanced sampling. The common belief behind these 2-stage methods is that the instance-balanced sampling learns better and more general representations, which has been demonstrated via better experimental results. However, the 2-stage methods fail to explain the essential reason for the poor performance in long-tailed data distribution. Moreover, these methods involve many tricks to obtain better results, which brings difficulties and complexities for training and poor interpretability. Some work attempts to solve the issue by balancing the gradients. \cite{tan2020equalization} filters out the gradients from major classes via an $0$-$1$ mask when back-propagating the gradients for samples of minor classes. However, besides the heavy reliance on many hyper-parameters, they only filter out gradients that are too large and still do not essentially solve the imbalance of the remaining gradients. Therefore, the key to solving the performance drop under the long-tailed distribution is to balance each gradient component from different classes.

A recent study \cite{papyan2020prevalence} observes the \textit{Neural Collapse} phenomenon. When the model converges to optimum in an ideal state, \ie, a balanced data distribution, four phenomenons can be observed:
\begin{enumerate}
	\item The intra-class features will collapse to their class means.
	\item The vectors of the class means will collapse to an ETF (Equiangular Tight Frame) geometric structure, \ie, the class mean vectors will have consistent length and equal-sized angles between any given pair, where the pairwise-distanced are maximized.
	\item The class means and the classifier weights will dually converge to each other so that the dot product of the features and the classifier weights will obtain the maximum for matched classes while the minimum for unmatched classes.
	\item The classifier decides the class according to the Euclidean distances among the feature vector and the classifier weights.
\end{enumerate}

\noindent However, under an imbalanced data distribution, the law of \textit{Neural Collapse} is no longer satisfied, and another unexpected phenomenon, \textit{Minority Collapse} \cite{fang2021exploring}, occurs, \ie, some weight vectors are close and even merged. The collapsed weights can no longer effectively distinguish features of different classes, which explains the deteriorated performance of classification on imbalanced data. We illustrate these two observations in \cref{fig:neural_minority_collapse}.

Inspired by the \textit{Neural Collapse} phenomenon, in this paper, we propose Attraction-Repulsion-Balanced Loss, abbreviated as ARB-Loss, to balance the gradients from different classes. We first analyze that the essential reason of \textit{Minority Collapse} is the imbalance between the \textit{attractive} and \textit{repulsive} gradient components from different classes. The classifier weights of minority classes receive a much stronger \textit{repulsion} gradients from majority-class features compared to the \textit{attractive} gradients from intra-class features. Therefore, to mitigate the \textit{Minority Collapse} dilemma, ARB-Loss adds coefficients on the denominator of the softmax function to balance the gradients from different classes. Compared to the state-of-the-art methods for imbalanced learning problems \cite{kang2019decoupling,zhong2021improving}, our ARB-Loss can achieve comparable or even better performance without bells and whistles.

The contributions of this paper can be summarized as follows:

\begin{enumerate}
	\item We analyze that the cause of \textit{Minority Collapse} is the imbalance between the \textit{attractive} and \textit{repulsive} gradient components from different classes. Based on the analyses, we propose ARB-Loss to balance the gradients from different classes to mitigate the \textit{Minority Collapse} dilemma under the imbalanced distribution.
	\item We analyze the properties of ARB-Loss. Both the theoretical and empirical results show that compared with the Cross-Entropy loss, ARB-Loss can achieve more balanced gradients and lead to a model closer to the neural collapse state.
	\item We perform experiments on large-scale long-tailed image classification and segmentation tasks. The imperial results suggest that our method can achieve comparable or even better performance without bells and whistles.
\end{enumerate}

In the following sections, we first review the related works of long-tailed learning and the studies of the \textit{Neural Collapse} phenomenon in Section \ref{sec:related_works}. Then, in Section \ref{sec:rethinking}, we rethink the learning mechanism under the traditional cross-entropy loss from the perspective of the gradients and analyze the cause of the unexpected \textit{Minority Collapse}. In Section \ref{sec:methods}, we elaborate on our ARB-Loss and provide relevant analytical results. In Section \ref{sec:experiments}, we perform experiments to show the effectiveness of our method. Finally, in Section \ref{sec:conclusion}, we make a conclusion for this paper and a prospective for future work.

\section{Related Works}\label{sec:related_works}

\subsection{Long-tailed Learning}

\textbf{Re-sampling.} Early works \cite{chawla2002smote,drumnond2003class,han2005borderline,buda2018systematic} re-sample the data (over-sample the samples of minor classes or under-sample the samples of major classes or both) to balance the data distribution. Over-sampling repeats the sample of minor classes, which might make the model overfit these classes and harm the generality. Under-sampling removes some samples of major classes, which might remove the key data for representation learning and bring the performance drop.

\textbf{Re-weighting.} Another idea is to assign different weights for different classes, even instances. \cite{huang2016learning,rota2017loss,cui2019class,tan2020equalization} re-weight the loss according to the class cardinalities. \cite{ren2020balanced} introduce the prior probabilities from a Bayesian view and balance the exponential logits via the class cardinalities. Although \cite{ren2020balanced} shares a similar formulation of the loss function with ours, we derive from a completely new perspective and offer in-depth analyses. Focal loss \cite{lin2017focal,mukhoti2020calibrating} re-weight each instances according to their \textit{hard} level, \ie, making the model to pay attention to the wrong-recognized samples. 

\textbf{Two-stage Learning.} An observation is that the representations learned under the instance-balancing sampling are more general. So many works try to split learning into two stages. \cite{cao2019learning}  first trains the model in a normal way in stage 1 and then uses deferred resampling to fine-tune with class-balanced resampling or uses deferred re-weighting to re-weight different classes in stage 2. \cite{zhou2020bbn} spatially disentangles the regular feature learning and re-balancing learning via two parallel branches.\cite{kang2019decoupling,zhong2021improving} decouple the representation learning and classification. They concluded that the instance-balanced sampling gives more general representations. They first used instance-balanced sampling to learn the representations at stage-1. Then, they freeze the feature and retrain the classifier (cRT) via techniques such as label-aware smooth (LAS) and learnable weight scaling (LWS).

\subsection{Neural Collapse}

A recent study \cite{papyan2020prevalence} discovers the classifier will converge to the state called \textit{Neural Collapse} when trained on the balanced data. Especially, \textit{Neural Collapse} is a phenomenon that the mean vectors of the intra-class features and the classifier weight vectors dually converge to an ETF (Equiangular Tight Frame) geometric structure, and the decision is based on the Euclidian distance among the feature vectors and the classifier weights. In the \cref{sec:nc}, we give a detailed description for the \textit{Neural Collapse} phenomenon. Most of existing theoretical methods \cite{mixon2020neural,weinan2020emergence,lu2020neural,zhu2021geometric,han2021neural,ji2021unconstrained} only analyse the reason of \textit{Neural Collapse} phenomenon when being trained under a balanced data distribution. However, for most real-world data, the class distribution is usually imbalanced. In the imbalanced cases, an unexpected phenomenon, \textit{Minority Collapse} \cite{fang2021exploring}, occurs. The classifier weights of minority classes will converge to similar directions. These theoretical works give an explainable view of the mechanism of the classifier under balanced or imbalanced distribution. However, there is no design inspired by \textit{Neural Collapse} for the long-tailed learning. Some recent work \cite{yang2022we} tries to avoid the gradient imbalance via fixing the classifier weights. However, it is not guaranteed that a fixed length of classifier weight does not harm representation learning. In this paper, we reveal that the \textit{Minority Collapse} phenomenon trained under imbalanced data distribution is essentially caused by the imbalanced gradient components. Inspired by this idea, we propose ARB-Loss from the \textit{Neural-Collapse} view to balance the \textit{attrative} and different \textit{repulsive} components of the gradients for the classifier weights, which are much more interpretable compared to previous methods.

\section{Rethinking the Cross Entropy Loss}\label{sec:rethinking}

\begin{figure}[th]
	\vskip 0.2in
	\begin{center}
		\centerline{\includegraphics[width=0.6\columnwidth]{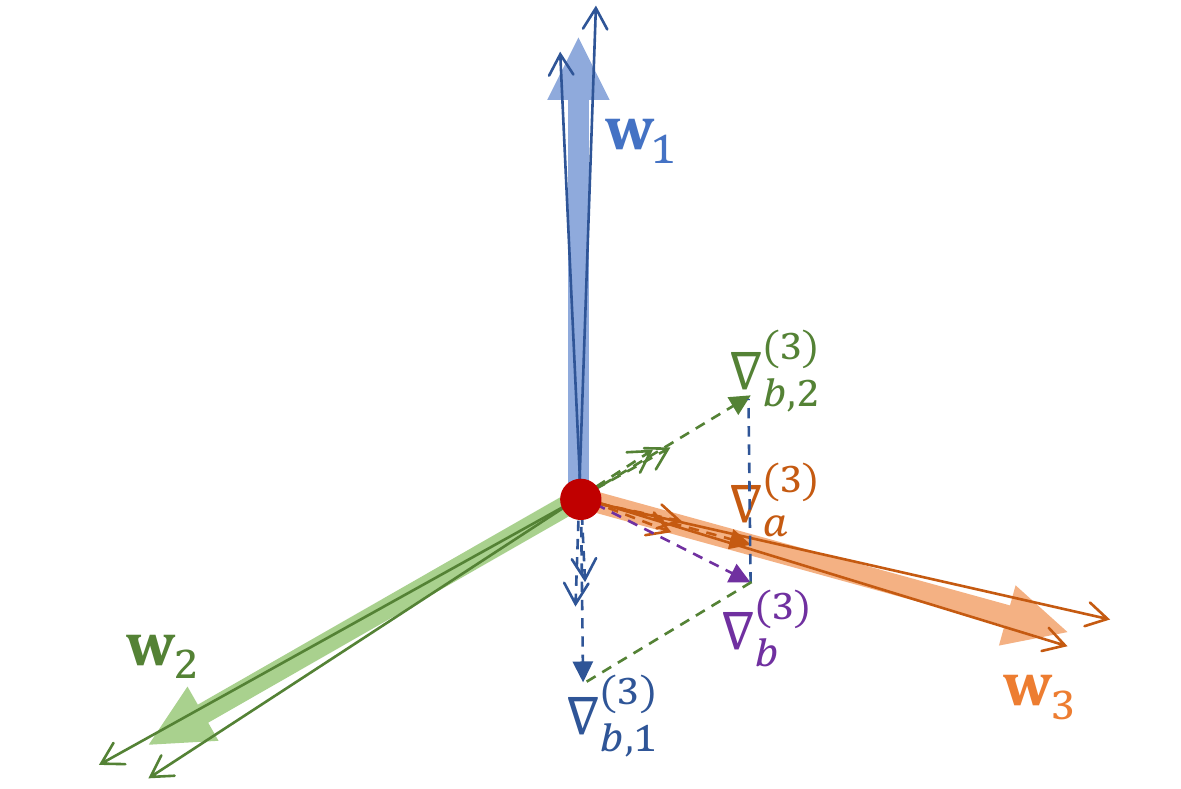}}
		\caption{\textbf{The \textit{attraction} and \textit{repulsion} components of the gradients.} For class $i$, the gradients from other classes (denoted as $\nabla^{(i)}_{b,j}, \; j \neq i$) act like \textit{repulsion force} to push the weights $\boldsymbol{w}_{i}$ to their inverse directions, while the gradients from class $i$ (denoted as $\nabla^{(i)}_{a}$) acts like a \textit{attraction force}. The imbalance among different components of $\nabla^{(i)}_{r}$ will cause the direction of $\boldsymbol{w}_{i}$ to incline to the inverse directions of major-class features. The imbalance between $\nabla^{(i)}_{a}$ and $\nabla^{(i)}_{r}$ will cause the norm of $\boldsymbol{w}_{i}$ to incline towards the direction of \textit{attraction} or \textit{repulsion}. (The thick solid arrows represent the classifier weights, the thin solid arrows represent the features, the thin dashed arrows represent the gradients, and the thin dashed arrows with the solid head represent the sum of gradients from each class. Each color represents a class, and the thin red dashed arrow represents the sum of all gradients.)}
		\label{fig:attraction_repulsion_grad}
	\end{center}
	\vskip -0.2in
\end{figure}

\begin{figure*}[ht]
	\vskip 0.2in
	\begin{center}
		\centerline{\includegraphics[width=\textwidth]{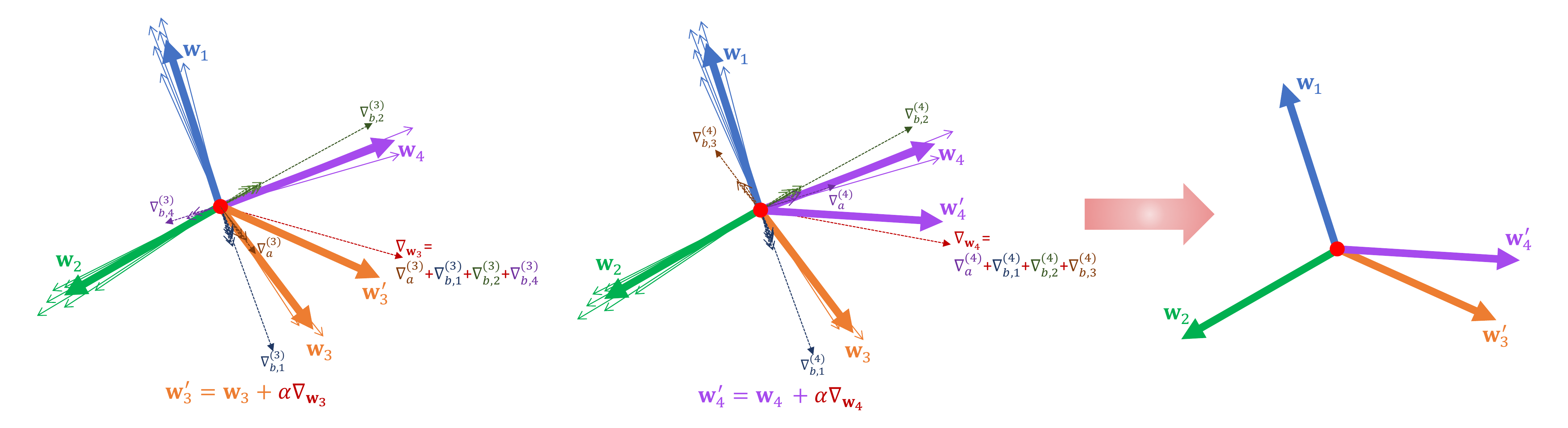}}
		\caption{\textbf{The cause of \textit{Minority Collapse} is the imbalance of gradients.} In this figure, we draw 4-class classification, where class $1$,$2$ (\textit{blue},\textit{green}) are major classes and $3$,$4$ (\textit{orange},\textit{purple}) are minor classes. We describe the updating process of the classifier weights of minor classes, \ie, $\boldsymbol{w}_{3}$, $\boldsymbol{w}_{4}$. For those minor classes, their gradients are mainly decided by the inverse direction of features of major classes. It seems that the major components of gradients have \textit{strong repulsion} to push the weights corresponding to minor classes to collapse together. (The color and arrow type have the same meaning as \cref{fig:attraction_repulsion_grad}.)}
		\label{fig:imbalanced_grad}
	\end{center}
	\vskip -0.2in
\end{figure*}

\textbf{Notation \& Problem Setup.} For a $c$-class classification task, we denote $\mathcal{X} = \{ \boldsymbol{x}_{i}, y_{i} \}, i \in \{ 1, \dots, n \}$ as the observed dataset, where $n$ is the number of samples, $\boldsymbol{x}_{i}$ and $y_{i} \in \{ 1, \dots, c \}$ are the $i$-th sample and its label respectively. We also use $\boldsymbol{y}_{i} \in \mathbb{R}^{c}$ as the corresponding vectorized label, e.g. the one-hot or smoothed one-hot label vector. Suppose that class $i$ has $n_{i}$ samples in the dataset $\mathcal{X}$, we use $\pi(i)$ to represent the set of sample indices that belongs to the class $i$, \ie, $\pi(i) = \{j: y_{j} = i\}$.

A vanilla $L$-layer deep neural network can be represented as: 

\begin{equation}
	\Phi_{\boldsymbol{\Theta}}(\boldsymbol{x}) = \boldsymbol{W}_{L}\sigma(\boldsymbol{W}_{L-1}\sigma(\cdots\sigma(\boldsymbol{W}_{1}\boldsymbol{x} + \boldsymbol{b}_{1}) + \cdots )+ \boldsymbol{b}_{L-1}),
\end{equation}

\noindent where $\boldsymbol{W_{L}}$ is the final classifier weight. We omit its bias for simplicity because bias can be incorporated into the weights by adding a dimension filled with $1$ on the features. The optimization target is:

\begin{equation}
	\min_{\boldsymbol{\Theta}} \sum_{i=1}^{c} \sum_{j \in \pi(i)} \mathcal{L} (\Phi_{\boldsymbol{\Theta}}(\boldsymbol{x}_{j}), \boldsymbol{y}_{j}) + \frac{\lambda}{2}{\lVert \boldsymbol{\Theta} \rVert}^{2},
\end{equation}

\noindent where $\boldsymbol{\Theta} = \{\boldsymbol{W}_{i}, \boldsymbol{b}_{i}\}_{i=1}^{L}$ are the learnable parameters, $\mathcal{L}$ is the loss function, and $\lambda$ is the weight decay parameter.

We use $\boldsymbol{W} = {[ \boldsymbol{w}_{1}, \dots, \boldsymbol{w}_{c} ]}^{\mathrm{T}} \in \mathbb{R}^{d \times c}$ to represent the classifier weights and $\boldsymbol{H} = [ \boldsymbol{h}_{1}, \dots, \boldsymbol{h}_{n} ] \in \mathbb{R}^{d \times n}$ to note the features of the last layer, \ie, $\boldsymbol{H} = \{ \boldsymbol{h}_{i} = \sigma(\boldsymbol{W}_{L-1}\sigma(\cdots\sigma(\boldsymbol{W}_{1}\boldsymbol{x}_{i} + \boldsymbol{b}_{1}) + \cdots )+ \boldsymbol{b}_{L-1}) \vert i = 1, 2, \dots, n \}$. Because the norms of the weights $\{\boldsymbol{W}_{i}\}_{i=1}^{L}$ and bias $\{\boldsymbol{b}_{i}\}_{i=1}^{L-1}$ have supremums and the popular activation functions also have finite supremums (e.g. Sigmoid, tanh etc.) or keep linear mapping in the non-restraint part (e.g. ReLU), the norm of $\boldsymbol{H}$ also have a supremum:

\begin{equation}
	\frac{1}{n}{\lVert \boldsymbol{H} \rVert}^{2}_{F} = \frac{1}{n}\sum_{i=1}^{c}\sum_{j \in \pi(i)}{\lVert \boldsymbol{h}_{j} \rVert}^{2}_{2} \leq E_{H},
\end{equation}

\noindent which also means that, for a certain class $i$, the features also have supremum:

\begin{equation}
	\frac{1}{n_{i}} \sum_{j \in \pi(i)}{\lVert \boldsymbol{h}_{j} \rVert}^{2} \leq E_{H,i}, \quad i = \{1, 2, \dots c \},
\end{equation}

\noindent where $E_{H} \in \mathbb{R}, E_{H, i} \in \mathbb{R}, i \in \{1, \dots, n\}$ are constants.

\subsection{Neural Collapse}\label{sec:nc}

\cite{papyan2020prevalence} observes that the feature and weight vectors of the classifier will dually converge to a special geometric structure under the balanced data distribution, and this phenomenon is called \textit{Neural Collapse}. \textit{Neural Collapse} have four manifestations: 

\begin{enumerate}
	\item The features will converge to their class mean.
	\item Under a balanced data distribution, the class mean vectors will collapse to an ETF (Equiangular Tight Frame) geometric structure, which can be represented as:
		\begin{equation}
			\boldsymbol{W}^{*} = \sqrt{\frac{c}{c - 1}} \boldsymbol{P}(\boldsymbol{I}_{c} - \frac{1}{c}\boldsymbol{1}_{c}\boldsymbol{1}_{c}^{\mathrm{T}}),
		\end{equation}
		\noindent where $\boldsymbol{P} \in \mathbb{R}^{d \times c}, (d \geq c)$ is a partial orthogonal matrix such that $\boldsymbol{P}^{\mathrm{T}}\boldsymbol{P} = \boldsymbol{I}_{c}$, $\boldsymbol{I}_{c}$ is the $c \times c$ identical matrix and $\boldsymbol{1}_{c} \in \mathbb{R}^{c \times 1}$ is a vector filled with $1$. For the classifier weights which has collapsed to the ETF structure, they meet:
		\begin{equation}
			\boldsymbol{w}_{i}^{*\mathrm{T}}\boldsymbol{w}^{*}_{j} = \begin{cases}
				1, \quad\quad\;\;\; i = j \\
				- \frac{1}{c-1}, \quad i \neq j \\
			\end{cases},
		\end{equation}
		\noindent where $\boldsymbol{w}^{*}_{i}$ is the $i$-the column of $\boldsymbol{W}^{*}$, \ie, the weights corresponding to class $i$. 
	\item The classifier weights will dually converge to the mean vectors of the corresponding class, \ie, the classifier weight vectors will collapse to the consistent ETF geometric structure.
	\item The decision of the classifier is based on the Euclidean distances of the feature vectors and each the classier weight, \ie, deciding the class whose mean vector is nearest to the current features.
\end{enumerate}

When the classifier weights and feature vectors obey these phenomenons of \textit{Neural Collapse}, the Fisher's discriminant ratio \cite{fisher1936use} is maximized, \ie, the optimal geometric structures for classification problem.

However, for long-tailed learning, \textit{Neural Collapse} no longer emerges. Instead, \textit{Minority Collapse} occurs \cite{fang2021exploring}, which describes the phenomenon that the classifier weight vectors of minority classes converge in a similar direction, \ie,

\begin{equation}
	\lim_{\frac{n_{m}}{n_{i}} \to \infty, \frac{n_{m}}{n_{i}} \to \infty} \boldsymbol{w}_{i} - \boldsymbol{w}_{j} = \boldsymbol{0}_{d}
\end{equation}

\noindent where $n_{m}$ is the cardinality of one of the majority classes. Due to the collapse of the minority-class weight vectors, the classifier fails to discriminate these classes correctly, which explains the fundamental reason for the poor performance under the long-tailed data distribution. In the following, we analyze the gradients derived from cross-entropy loss and reveal that the cause of \textit{Minority Collapse} is the imbalance among gradient components from different classes.

\subsection{The Gradients of Cross-Entropy Loss}

The cross-entropy loss can be defined as:

\begin{equation}
	\begin{aligned}
		\mathcal{L}_{\mathrm{ce}} & = -\sum_{j=1}^{b}\sum_{i=1}^{c} \boldsymbol{y}_{j,i}\sigma(\boldsymbol{z}_{j,i}) \\
								  & = -\sum_{j=1}^{b}\sum_{i=1}^{c}\boldsymbol{y}_{j,i}\log\left(\frac{\exp(\boldsymbol{w}_{i}^{\mathrm{T}}\boldsymbol{h}_{j})}{\sum_{k=1}^{c}\exp(\boldsymbol{w}_{k}^{\mathrm{T}}\boldsymbol{h}_{j})}\right),
	\end{aligned}
\end{equation}

\noindent where $b$ represents the batch size, $\sigma$ is the softmax function, $\boldsymbol{z}_{j,i} = \boldsymbol{w}_{i}^{\mathrm{T}}\boldsymbol{h}_{j}$ is the logit of the $j$-th sample for the class $i$. When the label is a one-hot vector, we can derive the gradient for the classifier weights as the following:

\begin{equation}
\label{equ:gradw_ce}
	\nabla_{\boldsymbol{w}_{i}} = \underbrace{\sum_{j \in \pi(i)}-\left(1 - p_{j,i}\right) \boldsymbol{h}_{j}}_{\nabla_{a}^{(i)}} + \underbrace{\underset{k \neq i}{\sum_{k=1}^{c}}\sum_{j \in \pi(k)} p_{j,i} \boldsymbol{h}_{j}}_{\nabla_{r}^{(i)}},
\end{equation}

\noindent where $p_{j,i} = \sigma(\boldsymbol{z}_{j,i})$ is the probability that sample $j$ belongs to class $i$, $\pi(i)$ represents the set of sample indices that belongs to the class $i$. We decouple the gradients $\nabla_{\boldsymbol{w}_{i}}$ as two parts: $\nabla_{a}^{(i)}$ and $\nabla_{r}^{(i)}$. When the classifier weight $\boldsymbol{w}_{i}$ updates, it steps along the negative gradient direction. As illustrated in \cref{fig:attraction_repulsion_grad}, $\nabla_{a}^{(i)}$ has the inverse direction of $\boldsymbol{h}_{j}$ and push $\boldsymbol{w}_{i}$ and $\boldsymbol{h}_{j}$ together, which acts like an \textit{attraction force} from $\boldsymbol{h}_{j}$; $\nabla_{r}^{(i)}$ has the same direction of $\boldsymbol{h}_{j}$ and push $\boldsymbol{w}_{i}$ and $\boldsymbol{h}_{j}$ away, which acts like a \textit{repulsion force} from $\boldsymbol{h}_{j}$.

\textbf{Gradients-Imbalance.} \cite{papyan2020prevalence} observes the \textit{Neural Collapse} phenomenon that the classifier weights and the last-layer features collapse to a $c$-simplex geometric structure, \ie, an equiangular tight frame (ETF), when the training data are balanced. It is reasonable and intuitive. However, for the class-imbalanced training set, \textit{Minority Collapse} phenomenon occurs \cite{fang2021exploring}, \ie, the ETF vertices of the minor classes collapse together. \textit{Minority Collapse} limits the performance of deep learning models, especially in the minor classes. \cref{equ:gradw_ce} reveals that the gradients are composed of the \textit{attraction} and \textit{repulsion} parts. To see the gradient imbalance more clearly, we perform further simplification as the following:

\begin{equation}
	\label{equ:gradw_ce_simplified}
	\begin{aligned}
		\nabla_{\boldsymbol{w}_{i}} & = \sum_{j \in \pi(i)}-q_{j,i} \boldsymbol{h}_{j} + \underset{k \neq i}{\sum_{k=1}^{c}}\sum_{j \in \pi(k)} q_{j,i} \boldsymbol{h}_{j} \\ 
								& = \underbrace{-n_{i} \bar{\boldsymbol{h}}_{(i)}}_{\nabla_{a}^{(i)}} + \underbrace{\sum_{k=1,k \neq i}^{c} n_{k}\bar{\boldsymbol{h}}_{(k)}}_{\nabla_{r}^{(i)}}, \\
	\end{aligned}
\end{equation}

\noindent where $n_{i}$ is the number of samples belonging to class $i$ and 

\begin{equation}
	\bar{\boldsymbol{h}}_{(i)} = \frac{1}{n_{i}}\underset{j \in \pi(i)}{\sum}q_{j,i}\boldsymbol{h}_{j}
\end{equation}

\noindent is a weighted feature mean of class $i$ and

\begin{equation}
	\label{equ:4}
	q_{j,i} = \begin{cases}
		1 - p_{j,i}, \quad j \in \pi(i)\\
		p_{j,i}, \quad\;\;\;\;\;\; j \notin \pi(i)
	\end{cases}
\end{equation}

\cref{equ:gradw_ce_simplified} shows that the coefficients of the gradients from samples of different classes depend on the class cardinalities. When there are huge disparities among the cardinality of major classes and minor classes, the gradient norms are imbalanced, and the classifier weights are most affected by the major classes. It causes the weights corresponding to minor classes to collapse due to the \textit{strong repulsion} effect from the features of major classes. For minority classes, the \textit{repulsion forces} that they receive have almost the same direction. So, after training, their classifier weight vectors will be pushed in a similar direction, even merging, which explains the reason for \textit{Minority Collapse}. We illustrate this issue in \cref{fig:imbalanced_grad}.

\section{Methods}\label{sec:methods}

\subsection{Balancing the Gradients}

Through the above analysis, the key to solving the performance drop when facing imbalanced class distribution is to balance the gradients from different class features. Our main idea to solve the gradients imbalance is to add balancing coefficients on each component of $\nabla_{\boldsymbol{w}_{i}}$. We hope $\nabla_{\boldsymbol{w}_{i}}$ has the following format:

\begin{equation}
	\label{equ:target_grad}
	\nabla_{\boldsymbol{w}_{i}} = -C\left(1 - p_{i}\right) \bar{\boldsymbol{h}}_{i} + \underbrace{C p_{i}\bar{\boldsymbol{h}}_{1} + \cdots + C p_{i}\bar{\boldsymbol{h}}_{c}}_{c-1},
\end{equation}

\noindent where $C$ is a positive constant. 

\subsection{ARB-Loss}

To make the gradients $\nabla_{\boldsymbol{w}_{i}}, \; i \in \{ 1, \dots, c \}$ satisfy the format of \cref{equ:target_grad}, we design a novel classification loss function. We name it \textit{Attraction-Repulsion-Balanced Loss}, abbreviated as ARB-Loss. ARB-Loss is formulated as follows:

\begin{equation}
\label{equ:arbloss}
	\mathcal{L}_{\mathrm{arb}} = -\sum_{j=1}^{b}\sum_{i=1}^{c}\boldsymbol{y}_{j,i}\log\left( \frac{\exp(\boldsymbol{w}_{i}^{\mathrm{T}}\boldsymbol{h}_{j})}{\sum_{k=1}^{c}\frac{n_{k}}{n_{i}}\exp(\boldsymbol{w}_{k}^{\mathrm{T}}\boldsymbol{h}_{j})} \right),
\end{equation}

\noindent where $n_{i}$ is the number of samples belonging to class $i$. Note that $n_{i}$ can represent the class cardinality in the whole training dataset or the mini-batch. The difference between ARB-Loss and traditional cross-entropy loss is that we add coefficients on the denominator of the softmax function.

After simple derivation, we can obtain the gradients for the classifier weights under our ARB-Loss. 

\begin{equation}
	\label{equ:grad_arbloss}
	\begin{aligned}
		\tilde{\nabla}\boldsymbol{w}_{i} & = \sum_{j \in \pi(i)}-\left(1 - \tilde{\sigma}_{i}(\boldsymbol{w}_{i}^{\mathrm{T}}\boldsymbol{h}_{j})\right) \boldsymbol{h}^{(j)} \\
							   & \;\;\;\; + \underset{k \neq i}{\sum_{k=1}^{c}}\sum_{j \notin \pi(k)}\frac{n_{i}}{n_{k}} \cdot \tilde{\sigma}_{k}(\boldsymbol{w}_{i}^{\mathrm{T}}\boldsymbol{h}_{j}) \boldsymbol{h}^{(j)} \\
							   		& = \sum_{j \in \pi(i)}-\tilde{q}^{(i)}_{j,i} \boldsymbol{h}^{(j)} + \underset{k \neq i}{\sum_{k=1}^{c}}\sum_{j \notin \pi(k)}\frac{n_{i}}{n_{k}} \tilde{q}^{(k)}_{j,i} \boldsymbol{h}^{(j)} \\
							   		& = -\underbrace{n_{i}\bar{\boldsymbol{h}}_{(i)}}_{\tilde{\nabla}_{a}^{(i)}} + \underbrace{\sum_{k=1,k \neq i}^{c} n_{i}\bar{\boldsymbol{h}}_{(k)}}_{\tilde{\nabla}_{r}^{(i)}}, \\
	\end{aligned}
\end{equation}

\noindent where 

\begin{equation}
	\label{equ:8}
	\tilde{q}^{(\alpha)}_{j,i} = \begin{cases}
		1 - \tilde{p}^{(\alpha)}_{j,i}, \quad j \in \pi(i)\\
		\tilde{p}^{(\alpha)}_{j,i}, \quad\;\;\;\;\;\; j \notin \pi(i)
	\end{cases},
\end{equation}

\noindent where $\tilde{p}^{(\alpha)}_{j,i} = \tilde{\sigma}_{\alpha}(\boldsymbol{w}_{i}^{\mathrm{T}}\boldsymbol{h}_{j}) = \frac{\exp(\boldsymbol{w}_{i}^{\mathrm{T}}\boldsymbol{h}_{j})}{\sum_{k=1}^{c}\frac{n_{k}}{n_{\alpha}}\exp(\boldsymbol{w}_{k}^{\mathrm{T}}\boldsymbol{h}_{j})}$ is the corresponding softmax function in our case. Note that the subscript $\alpha$ of $\tilde{\sigma}_{\alpha}$ identifies the ground-truth class of the sample $j$. From \cref{equ:grad_arbloss}, we can see that the \textit{attraction} component $\tilde{\nabla}_{a}^{(i)}$ and each parts of the \textit{repulsion} component $\tilde{\nabla}_{r}^{(i)}$ have the same coefficients, \ie, attraction and repulsion terms are balanced.

\subsection{Analytical Results}

In this section, we further explore the properties of our ARB-Loss to show its superiority over the Cross-Entropy loss when training on imbalanced data. 

We introduce two propositions to demonstrate that ARB-Loss can efficiently reduce the imbalance of the gradients from different classes.

\begin{proposition}
	\label{prop:1}
	For gradient $\nabla_{\boldsymbol{w}_{i}}$ in cross entropy and $\tilde{\nabla}_{\boldsymbol{w}_{i}}$ in ARB-Loss, the proportion of the supremums of different repulsion terms' norm, $\frac{\sup \lVert \nabla_{b,k} \rVert}{\sup \lVert \nabla_{b,l} \rVert}$ and $\frac{\sup \lVert \tilde{\nabla}_{b,k} \rVert}{\sup \lVert \tilde{\nabla}_{b,l} \rVert}$ (where $k \neq l, k \neq i, l \neq i$), obey:
	\begin{equation}
		\begin{aligned}
			\text{CE Loss:   } \frac{\sup \lVert \nabla_{b,k} \rVert}{\sup \lVert \nabla_{b,l} \rVert} & \leq \frac{n_{k}}{n_{l}} \cdot C_{kl} \cdot \sqrt{\frac{E_{H,k}}{E_{H,l}}}, \\
			\text{ARB-Loss:  } \frac{\sup \lVert \tilde{\nabla}_{b,k} \rVert}{\sup \lVert \tilde{\nabla}_{b,l} \rVert} & \leq \tilde{C}_{kl} \cdot \sqrt{\frac{E_{H,k}}{E_{H,l}}}, 
		\end{aligned}
	\end{equation}
	\noindent where $C_{kl}, \tilde{C}_{kl} \in \mathbb{R}$ are constants.
\end{proposition}

\begin{proposition}
	\label{prop:2}
	For gradient $\nabla_{\boldsymbol{w}_{i}}$ in cross entropy and $\tilde{\nabla}_{\boldsymbol{w}_{i}}$ in ARB-Loss, the proportions of the supremums of the attraction and repulsion terms' norm, $\frac{\sup {\lVert \nabla_{a}^{(i)} \rVert}}{\sup {\lVert \nabla_{b,k}^{(i)} \rVert}}$ and $\frac{\sup {\lVert \tilde{\nabla}_{a}^{(i)} \rVert}}{\sup {\lVert \tilde{\nabla}_{b,k}^{(i)} \rVert}}$ (where $k \neq i$), obey:
	\begin{equation}
		\begin{aligned}
			\text{CE Loss:   } \frac{\sup {\lVert \nabla_{a}^{(i)} \rVert}}{\sup {\lVert \nabla_{b,k} \rVert}} & \leq \frac{n_{i}}{n_{k}} \cdot C_{ik} \cdot \sqrt{\frac{E_{H,i}}{E_{H,k}}}, \\
			\text{ARB-Loss:  } \frac{\sup {\lVert \tilde{\nabla}_{a}^{(i)} \rVert}}{\sup {\lVert \tilde{\nabla}_{b,k} \rVert}} & \leq \tilde{C}_{ik} \cdot \sqrt{\frac{E_{H,i}}{E_{H,k}}}, \\
		\end{aligned}
	\end{equation}
	\noindent where $C_{ik}, \tilde{C}_{ik} \in \mathbb{R}$ are constants.
\end{proposition}

\cref{prop:1} shows that ARB-Loss can reduce the imbalance of the supremum of different \textit{repulsion} gradient terms. The balanced \textit{repulsion} gradients can remedy the issue that the classifier weights of minority classes are pushed to the inverse of the majority class weights and collapse together. \cref{prop:2} shows that ARB-Loss can reduce the imbalance between the \textit{attraction} and arbitrary \textit{repulsion} terms. It is important if the current class $i$ is a minority class. If they are imbalanced in this case, the \textit{attraction force} is much less than the \textit{repulsion force} from majority classes and the \textit{attraction} is not enough to fight against the strong \textit{repulsion} that pushes the minority class features together.

Besides balancing the \textit{attraction} and \textit{repulsion} components of the gradients, compared to previous loss functions designed specifically for long-tailed learning, our ARB-Loss has the following advantages: \
\begin{enumerate}
	\item ARB-Loss can achieve comparable or even better results via only one-stage training, compared to previous two-stage methods \cite{cao2019learning,kang2019decoupling,zhong2021improving};
	\item ARB-Loss has no hyperparameters to adjust according to different data sources like \cite{lin2017focal,cui2019class,cao2019learning,tan2020equalization}.
\end{enumerate}

\section{Experiments}\label{sec:experiments}

In this section, we perform experiments on classification and segmentation tasks to analyze our ARB-Loss. We put the description of datasets used in this section and the implementation details in the \cref{appendix:dataset} and \cref{appendix:impl}.

\subsection{Empirical Results}\label{sec:vis_ana}

\subsubsection{The Geometric Structure of the Classifier Weights.}

We propose three balance metrics to measure \textit{how balanced the classifier weights are}, \ie, \textit{how similar the classifier weights are with the ideal ETF geometric structure described by the Neural Collapse phenomenon.} 

\begin{equation}
	\begin{aligned}
		\mathcal{B}_{\mathcal{D}}^{2}(\boldsymbol{W}) &= \frac{1}{(c-1)^{2}}\sum_{i=1}^{c}\sum_{j=1,j \neq i}^{c} \left( \boldsymbol{w}_{i}^{\mathrm{T}}\boldsymbol{w}_{j} - \frac{1}{(c-1)^{2}}\sum_{i=1}^{c}\sum_{j=1,j \neq i}^{c} \boldsymbol{w}_{i}^{\mathrm{T}}\boldsymbol{w}_{j} \right)^{2} \\
		\mathcal{B}_{\mathcal{A}}^{2}(\boldsymbol{W}) &= \frac{1}{(c-1)^{2}}\sum_{i=1}^{c}\sum_{j=1,j \neq i}^{c} \left( \langle \boldsymbol{w}_{i}, \boldsymbol{w}_{j} \rangle - \frac{1}{(c-1)^{2}}\sum_{i=1}^{c}\sum_{j=1,j \neq i}^{c} \langle \boldsymbol{w}_{i}, \boldsymbol{w}_{j} \rangle \right)^{2} \\
		\mathcal{B}_{\mathcal{L}}^{2}(\boldsymbol{W}) &= \frac{1}{c} \sum_{i=1}^{c} \left( {\lVert \boldsymbol{w}_{i} \rVert}_{2} - \frac{1}{c} \sum_{i=1}^{c} {\lVert \boldsymbol{w}_{i} \rVert}_{2} \right)^{2} \\
	\end{aligned}
\end{equation}

\noindent where $\langle \boldsymbol{w}_{i}, \boldsymbol{w}_{j} \rangle = \frac{\boldsymbol{w}_{i}^{\mathrm{T}}\boldsymbol{w}_{j}}{{\lVert \boldsymbol{w}_{i} \rVert}_{2}\cdot{\lVert \boldsymbol{w}_{j} \rVert}_{2}}$ is the cosine of the angle between $\boldsymbol{w}_{i}$ and $\boldsymbol{w}_{j}$. $\mathcal{B}_{\mathcal{A}}^{2}(\boldsymbol{W})$ measures the balance degree of the angles of the weight vector pairs, $\mathcal{B}_{\mathcal{L}}^{2}(\boldsymbol{W})$ measures the variance of the norm of the weight vectors, and $\mathcal{B}_{\mathcal{D}}^{2}(\boldsymbol{W})$ measures the balance degree of the dot-product of the weight vector pairs. In the ideal case, \ie, the ETF geometric structure, the angles of all weight vector pairs, and the norms of all weight vectors are equivalent.

\begin{figure*}[ht]
	\vskip 0.2in
	\begin{center}
		\centerline{\includegraphics[width=\textwidth]{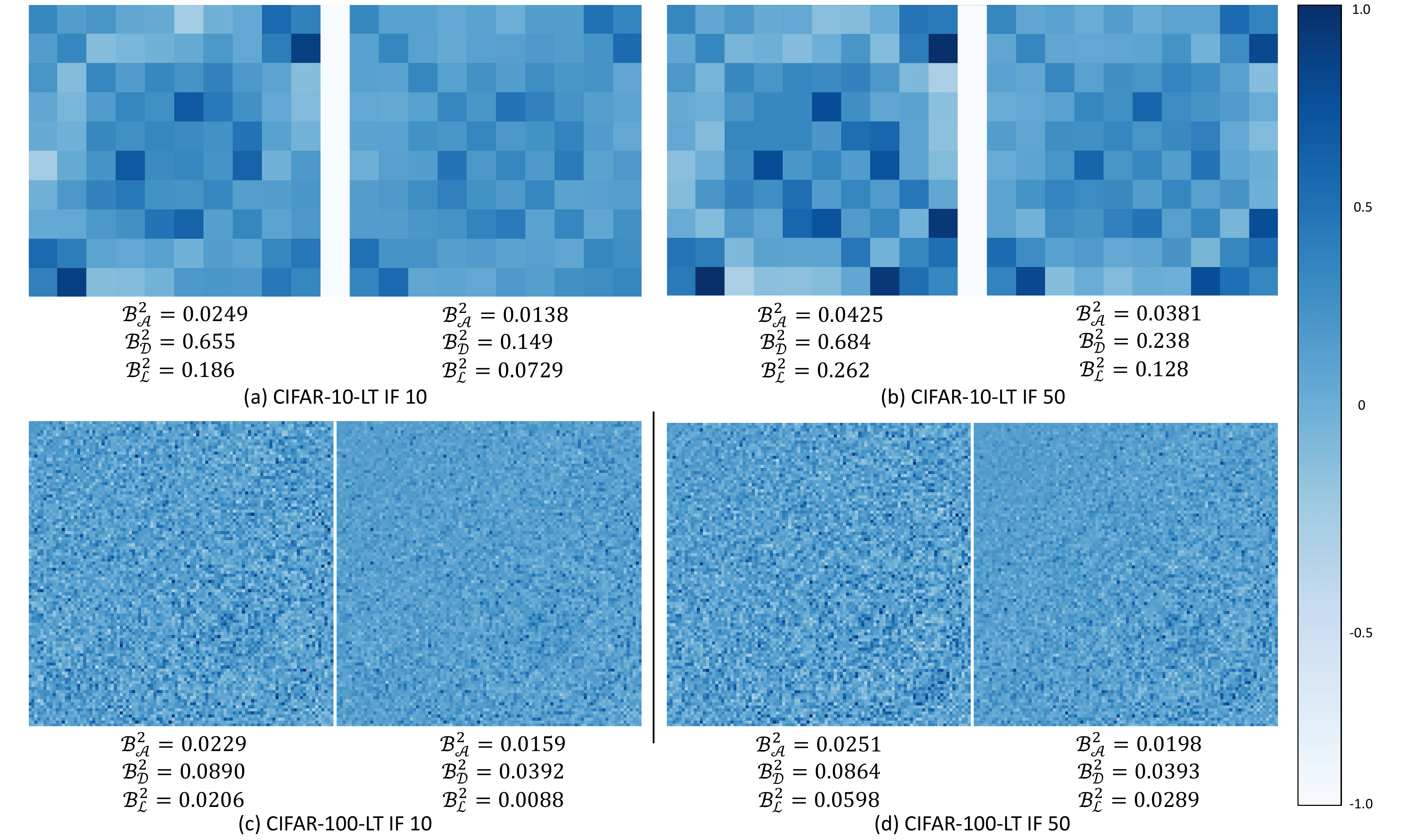}}
		\caption{\textbf{The geometric structure of the classifier weights}. In the (a),(b),(c), and (d) 4 sub-figures, the left part represents the classifier trained with cross-entropy loss, and the right part represents the classifier trained with ARB-Loss. The visualization details and meanings are elaborated in \cref{sec:vis_ana}. The color variance of the left part is larger than the right part, which demonstrates that the classifier weights trained by ARB-Loss are more dispersed than the ones trained by cross-entropy loss.}
		\label{fig:vis_angle}
	\end{center}
	\vskip -0.2in
\end{figure*}

For better understanding, we visualize the similarity of the classifier weights in \cref{fig:vis_angle}. In the figure, each sub-figures represents a matrix $\boldsymbol{M} \in \mathbb{R}^{c \times c}$ whose $i,j$-th entry is:

\begin{equation}
	\boldsymbol{M}_{i,j} = \begin{cases}
		\frac{\boldsymbol{w}_{i}^{\mathrm{T}}\boldsymbol{w}_{j}}{\sum_{j=1,j \neq i}^{c}\boldsymbol{w}_{i}^{\mathrm{T}}\boldsymbol{w}_{j} }, \quad i \neq j \\
		0, \quad \quad \quad \quad \quad \quad \; \; i = j
	\end{cases},
\end{equation}

\noindent Because the diagonal of matrix $\boldsymbol{M}$ is meaningless, we set them to zero. Below each sub-figure, we also indicate the values of the three balance metrics mentioned above. In the ideal state, we expect the variance of each row (except for the diagonal) of $\boldsymbol{M}$ to be zeros: 

\begin{equation}
	\boldsymbol{M}_{i,j} = \frac{1}{c-1}, \quad i,j \in \{ 1, \dots, c \}, i \neq j, .
\end{equation}

\noindent The ideal values of the balance metrics are:

\begin{equation}
	\mathcal{B}_{\mathcal{D}}^{*2}(\boldsymbol{W}) = \mathcal{B}_{\mathcal{A}}^{*2}(\boldsymbol{W}) = \mathcal{B}_{\mathcal{L}}^{*2}(\boldsymbol{W}) = 0 \\
\end{equation}

In the figure, we can see that the color variance of the left part (light and dark colors) is larger than the right part (all dark colors), which demonstrates that the classifier weights trained by ARB-Loss are more dispersed than the ones trained by cross-entropy loss. ARB-Loss can remarkably reduce the variance of each row of $\boldsymbol{M}$, which means that our ARB-Loss can mitigate the \textit{Minority Collapse} dilemma. 

\subsubsection{Different Components of Gradients.}

\begin{figure}[t]
	\vskip 0.2in
	\begin{center}
		\centerline{\includegraphics[width=\columnwidth]{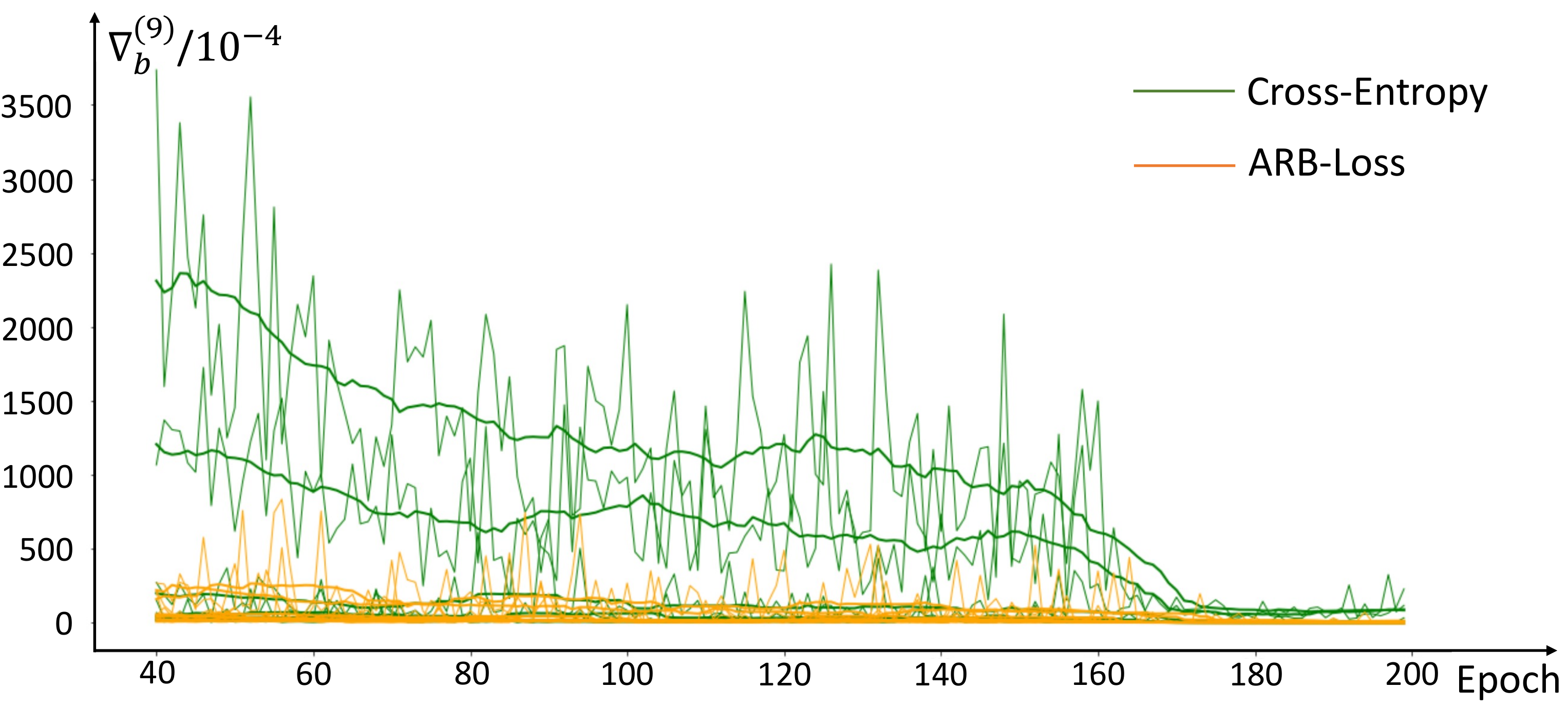}}
		\caption{\textbf{Visualzation example of the different components of the gradients.} In the figure, we take the training on CIFAR-10-LT dataset as an example. We plot the curve of magnitude (\ie, the vector length) of the different components $\nabla^{(9)}_{b,i}, \; i \neq 9$ of the gradients $\nabla_{\boldsymbol{w}_{9}}$, where $\boldsymbol{w}_{9}$ the classifier weights for class $9$ whose cardinality is the least. The thin and thick lines represent the original and smoothed curves, respectively. We show the results from the $40$-th epoch to ignore the instability in the initial training.}
		\label{fig:vis_grad_prop}
	\end{center}
	\vskip -0.2in
\end{figure}

We show a visualization example in \cref{fig:vis_grad_prop}. Each point on the original curves (the \textit{thin} line in the figure) represents the mean lengths of the gradients in one epoch, \ie,

\begin{equation}
	g_{k} = \frac{1}{N / B}\sum_{n=0}^{N/B} \lVert \nabla_{b,k}^{(9)} \rVert = \frac{1}{N / B}\sum_{n=0}^{N/B}\left\Vert \sum_{j\in\pi(k)}^{} \frac{\partial \mathcal{L}^{(j)}}{\boldsymbol{w}_{k}} \right\Vert,
\end{equation}

\noindent where $N$ is the total number of samples, $B$ is the batch size, $\frac{\mathcal{L}^{(j)}}{\boldsymbol{w}_{k}}$ is the gradients of the loss on $j$-th sample with respect to $\boldsymbol{w}_{k}$. From the visualization, we can see that using cross-entropy loss on imbalanced data will introduce huge magnitude differences among different gradient components, and using our proposed ARB-Loss can balance each gradient component to make them have a similar magnitude.

\subsection{Results on Image Classification Task}

Experiment results are shown in \cref{tab:exp_cifar} and \cref{tab:exp_largescale}. For previous methods, we directly copy the results from the original papers, except for the results noted with $\dagger$. 

\begin{table*}[th]
\centering
\resizebox{\textwidth}{!}{
\begin{tabular}{l|c|cccc|cccc}
\hline
\multirow{2}{*}{Methods} & \multirow{2}{*}{stage} & \multicolumn{4}{c|}{CIFAR-10(-LT)} & \multicolumn{4}{c}{CIFAR-100(-LT)} \\ \cline{3-10} 
                         &                        & 100    & 50    & 10    & Normal  & 100    & 50    & 10    & Normal  \\ \hline
Plain Model \cite{he2016deep} & 1                      & 70.4   & 74.8  & 86.4  & $92.8^{\dagger}$        & 38.4   & 43.9  & 55.8  &   $68.7^{\dagger}$      \\
mixup \cite{zhang2018mixup} & 1                      & 73.1   & 77.8  & 87.1  &         $92.0^{\dagger}$ & 39.6   & 45.0  & 58.2  & $68.7^{\dagger}$      \\
LDAM+DRW \cite{cao2019learning} & 1 & 74.9 & - & 86.7 & - & 40.3 & - & 57.3 & - \\
Remix+DRW \cite{chou2020remix} & 2  & 79.8 & - & 89.1 & - & 46.8 & - & 61.3 & - \\
BBN \cite{zhou2020bbn} & 2                      & 79.9   & 82.2  & 88.4  &  -       & 42.6   & 47.1  & 59.2  &  -       \\
MiSLAS \cite{zhong2021improving} & 2                      & 82.1   & 85.7  & 90.0  &    -     & 47.0   & 52.3  & 63.2  &   -      \\ \hline
ARB-Loss                 & 1                      & 83.3   & 85.7  & 90.2  & 92.6       & 47.2   & 52.6  & 62.1  & 68.6        \\
\hline
\end{tabular}
}
\caption{Top-1 accuracy (\%) on CIFAR-10(-LT) and CIFAR-100(-LT). The experiments are performed on ResNet-32 \cite{he2016deep}. In the table head, $100$, $50$, $10$ represents $\mathrm{IF} = 100,50,10$.\textit{Normal} represents the original dataset without undersampling. $\dagger$ represents our reimplement results.}
\label{tab:exp_cifar}
\end{table*}

\begin{table*}[th]
\centering
\resizebox{\textwidth}{!}{
\begin{tabular}{l|c|cccc|cccc|cccc}
\hline
\multirow{2}{*}{Methods} & \multirow{2}{*}{Stage} & \multicolumn{4}{c|}{ImageNet-LT} & \multicolumn{4}{c|}{Places-LT} & \multicolumn{4}{c}{iNaturalist2018} \\ \cline{3-14} 
                         &                        & Many  & Median  & Few  & Overall & Many  & Median  & Few  & Overall & Many  & Median  & Few  & Overall  \\ \hline
Plain Model \cite{he2016deep} & 1                      & $65.5^{\dagger}$ & $40.3^{\dagger}$ & $13.1^{\dagger}$ & $45.6^{\dagger}$ &  -   & -      & -    & -     & $72.8^{\dagger}$ & $63.7^{\dagger}$ & $58.4^{\dagger}$ & $62.3^{\dagger}$     \\
CB-Focal \cite{cui2019class} & 1                      & -     & -       & -    &    -    & -    & -      & -    &  -   & -      & -       & -     & 61.1     \\
RangeLoss \cite{zhang2017range} & 1&- &- &- &- & 41.1 & 35.4 & 23.2 & 35.1 & -& -& -&- \\
Focal+DRW \cite{lin2017focal} &1 & -&- &- & 47.9 &- &- & -& -& -& -& -&- \\ 
CE+DRW \cite{cao2019learning} &    1                    &  -     &    -     &      -& 48.5    &  -    &   -     &   -   &   -      &    -    &     -    &  -     &    -      \\
LDAM+DRW \cite{cao2019learning} &    1                    &  -     &    -     &    -  & 48.8    &   -   &   -     &  -    &   -      &   -     &    -     &   -    & 68.0     \\
OLTR \cite{liu2019large} &    1                    & 43.2  &  35.1  & 18.5  & 35.6 & 44.7 & 37.0   & 25.3 & 35.9    &    -    &    -     &   -    & -         \\
OTLR+LFME \cite{xiang2020learning} &1 & 47.0 & 37.9 & 19.2 & 38.8 & 39.3 & 39.6 & 24.2 & 36.2 &- &- &- &- \\
Remix+DRW \cite{chou2020remix} & 2 & - & - & - & - & -& -&- & -& -& -&- & 70.5 \\
BBN \cite{zhou2020bbn} & 2                     &   -    &     -    &  -    &   -      &   -   &     -   &  -    &     -    &      -  &      -   &      - & 69.6     \\
cRT+mixup \cite{zhong2021improving}  &       2                 & 63.9  & 49.1    & 30.2 & 51.7    & 44.1 & 38.5   & 27.1 & 38.3    & 74.2   & 71.1    & 68.2  & 70.2     \\
LWS+mixup \cite{zhong2021improving} &  2                      & 62.9  & 49.8    & 31.6 & 52.0    & 41.7 & 41.3   & 33.1 & 39.7    & 72.8   & 71.6    & 69.8  & 70.9     \\
MiSLAS\cite{zhong2021improving} & 2                      & 61.7  & 51.3    & 35.8 & 52.7    & 39.6 & 43.3   & 36.1 & 40.4    & 73.2   & 72.4    & 70.4  & 71.6     \\ \hline
ARB-Loss                 & 1                      & 60.2  & 51.8    & 38.3 & 52.8    & 41.9 & 41.5   & 32.1 & 39.7    & 71.9   & 72.1    & 71.7  & 71.7     \\ \hline
\end{tabular}
}
\caption{Top-1 accuracy (\%) on ImageNet-LT, Places-LT and iNaturalist2018. The experiments of ImageNet-LT and iNaturalist2018 are performed on ResNet-50, and the ones of Places-LT are performed on ResNet-152. $\dagger$ represents our reimplement results. }
\label{tab:exp_largescale}
\end{table*}

\textbf{CIFAR-10(-LT) \& CIFAR-100(-LT).} \cref{tab:exp_cifar} shows the results on CIFAR-10/100-LT. When performing experiments on CIFAR-10/100-LT, the class cardinalities used in ARB-Loss, are collect in a batch, \ie, the $n_{i}, \; i \in \{1, \dots, c\}$ in \cref{equ:arbloss} are dynamic for each different batches. Therefore, when the dataset is balanced, ARB-Loss \textit{approximately} degrades as the cross-entropy loss. The reason why it is said \textit{approximate} is that the samples are not balanced in a batch, although the whole training set is balanced. Therefore, we compare ARB-Loss with the plain model (using traditional cross-entropy) in the \textit{Normal} column. From the results, although the improvements achieved by our ARB-Loss are slight, it is also significant, considering that our method only needs one-stage training while recent state-of-the-art long-tailed works need 2-stage training. Moreover, the results on \textit{Normal} case suggest that our ARB-Loss is approximate to the traditional cross-entropy loss, although ARB-Loss is dynamic for every mini-batch.

\textbf{Large-scale Image Classification Datasets.} We list the results on the ImageNet-LT, Places-LT and iNaturalist2018 in \cref{tab:exp_largescale}. On the large-scale datasets, we found that using the global class cardinalities (\ie, the class cardinalities in the whole training set) can obtain better performance than using dynamic class cardinalities for each batch. Therefore, for the ARB-Loss experiments in \cref{tab:exp_largescale}, we all use the global class cardinalities. The results suggest that our ARB-Loss can achieve comparable results with current state-of-the-art methods, e.g. MiSLAS \cite{zhong2021improving}. It is worth noting that the results of ARB-Loss are obtained from only one-stage training, and the nowadays SOTA long-tail methods such as. BBN \cite{zhou2020bbn}, MiSLAS \cite{zhong2021improving}, need training for 2 stages.

\subsection{Results on Segmentation Task}

\begin{table}[t]
\centering
\begin{tabular}{c|cc|cc}
\hline
Models      & \multicolumn{2}{c|}{U-Net} & \multicolumn{2}{c}{Deeplabv3+} \\ \hline
Metrics     & mIoU   & mAcc   & mIoU      & mAcc      \\ \hline
baseline & $67.6^{\dagger}$ & $74.6^{\dagger}$ & $79.3^{\dagger}$ & $86.3^{\dagger}$   \\
ARB-Loss & 67.9 & 79.7 & 79.5 & 89.0 \\ \hline
\end{tabular}
\caption{The Segmentation Results on CityScapes dataset. \textit{mAcc} is the mean of per-class pixel accuracy and \textit{mIoU} is the mean of per-class IoU (Intersection over Union). $\dagger$ represents our reimplement results.}
\label{tab:exp_seg}
\end{table} 

We perform image segmentation experiments on two baselines: U-Net \cite{ronneberger2015u} and Deeplabv3+ \cite{chen2018encoder}. The results are shown in \cref{tab:exp_seg}. Our ARB-Loss can significantly improve pixel accuracy. The reason why the improvements on \textit{mIoU} metric are slight might be that the denominator of IoU involved FP (\textit{false positive}), compared to the denominator of pixel accuracy. In \cref{tab:detail_seg}, we show the per-class results. We can see that the classes whose IoU gets improved while the pixel accuracy is still a little lower are almost the minor classes. If the FP item is relatively too large for the cardinality of a minor class, the IoU value will drop. Meanwhile, if these FP belongs to some major classes, it will have little effect on the value of pixel accuracy. 

\section{Conclusion \& Future Works}\label{sec:conclusion}

In this paper, we analyze that the reason for the performance drop under long-tailed distributions is the imbalances of the gradients from different classes. Then, we propose Attraction-Repulsion-Balanced Loss (ARB-Loss) to balance the different gradient components. Moreover, we analyze the properties of our ARB-Loss to theoretically demonstrate the rationality of our design. We also give some visualization examples to analyze the geometric structure of the classifier weights and the magnitude proportions of the gradient components. Moreover, we perform experiments on large-scale classification and segmentation datasets to demonstrate the effectiveness of ARB-Loss. ARB-Loss can achieve state-of-the-art performance via only one-stage training. 

For the future, the following topics are worth studying:

\begin{itemize}
	\item The geometric structure described by \textit{Neural Collapse} also has great inspiration for the bottlenecks of other machine-learning tasks, \ie, few-shot or online learning \etc How to design algorithms inspired by \textit{Neural Collapse} for the other tasks is a topic worth future research.
	\item Besides the classifier, the optimization of the features extracted by the backbone networks is also affected by the imbalanced distribution. How to solve this issue is still an open question.
\end{itemize}

\noindent In the future, we will continue the relevant research topics along the line of \textit{Neural Collapse} and explore 
the underlying principles behind it.

\section{Acknowledgements}

This work was supported in part by The National Key Research and Development Program of China (Grant Nos: 2018AAA0101400), in part by The National Nature Science Foundation of China (Grant Nos: 62036009, U1909203, 61936006, 62133013), in part by Innovation Capability Support Program of Shaanxi (Program No. 2021TD-05).

\bibliography{mybibfile}

\newpage
\appendix
\onecolumn
\section{Proofs}

\subsection{Proof of \cref{prop:1}}

\begin{proof}
	
	For $k$-th($k \neq i$) \textit{repulsion} terms, 
	
	\begin{equation}
		\begin{aligned}
		\lVert \nabla_{b,k} \rVert & = \lVert \sum_{j \in \pi(k)}p_{j,i} \boldsymbol{h}_{j} \rVert \\
			& \leq \sum_{j \in \pi(k)}p_{j,i} \lVert \boldsymbol{h}_{j} \rVert \\
			& \leq \sqrt{\sum_{j \in \pi(k)}p_{j,i}^{2}} \sqrt{\sum_{j \in \pi(k)}{\lVert \boldsymbol{h}_{j} \rVert}^{2}} \\
			& = B_{k}\sqrt{n_{k}{\overline{\lVert \boldsymbol{h}_{(k)} \rVert}}^{2}} \leq B_{k}\sqrt{n_{k}E_{H,k}}
		\end{aligned},
	\end{equation}
	
	\noindent where
	
	\begin{equation}
		B_{k} = \sqrt{\sum_{j \in \pi(k)}p_{j,i}^{2}}  .
	\end{equation}
	
	\noindent and
	
	\begin{equation}	
		\overline{{\lVert \boldsymbol{h}_{(i)} \rVert}^{2}} = \frac{\sum_{j \in \pi(i)}{\lVert \boldsymbol{h}_{j} \rVert}^{2}}{n_{i}} \leq E_{H,i}.
	\end{equation}
	
	\noindent Note that we use the subscript with a bracket to represent the mean of the squared norm of some certain class features. The brackets are used to be distinguished from the feature indices. 
	
	For $j \in \pi(k), \; k \neq i$, there is a non-zero minimum for $p_{j,i}$ which is denoted as $p_{\mathrm{min}}^{(k,i)}$ and a maximum for $p_{j,i}$ which is less than $1$ and is denoted as $p_{\mathrm{max}}^{(k,i)}$. Then, 
	
	\begin{equation}
	\sqrt{n_{k}}p_{\mathrm{min}}^{(k,i)} \leq B_{k} \leq \sqrt{n_{k}}p_{\mathrm{max}}^{(k,i)}
	\end{equation}
	
	\noindent Therefore,
	
	\begin{equation}
		\frac{\sup {\lVert \nabla_{b,k} \rVert}}{\sup {\lVert \nabla_{b,l} \rVert}} = \frac{B_{k}\sqrt{n_{k}E_{H,k}}}{B_{l}\sqrt{n_{l}E_{H,l}}} \leq \frac{n_{k}}{n_{l}}\cdot C_{kl} \cdot \sqrt{\frac{E_{H,k}}{E_{H,l}}}
	\end{equation}
	
	\noindent where $C_{kl} = p_{\mathrm{max}}^{(k,i)}/p_{\mathrm{min}}^{(k,i)}$ is a constant.

	Similarly, for the ARB-Loss, 
	
	\begin{equation}
		\begin{aligned}
		\lVert \tilde{\nabla}_{b,k} \rVert & = \lVert \frac{n_{i}}{n_{k}}\sum_{j \in \pi(k)}\tilde{p}_{j,i} \boldsymbol{h}_{j} \rVert \\
			& \leq \frac{n_{i}}{n_{k}}\sum_{j \in \pi(k)}\tilde{p}_{j,i} \lVert \boldsymbol{h}_{j} \rVert \\
			& \leq \frac{n_{i}}{n_{k}}\sqrt{\sum_{j \in \pi(k)}\tilde{p}_{j,i}^{2}} \sqrt{\sum_{j \in \pi(k)}{\lVert \boldsymbol{h}_{j} \rVert}^{2}} \\
			& = \frac{n_{i}}{n_{k}}\cdot\tilde{B}_{k}\sqrt{n_{k}{\overline{\lVert \boldsymbol{h}_{(k)} \rVert}}^{2}} \leq \frac{n_{i}}{n_{k}}\cdot\tilde{B}_{k}\sqrt{n_{k}E_{H,k}}
		\end{aligned},
	\end{equation}
	
	\noindent where, 
	
	\begin{equation}
		\tilde{p}_{\mathrm{min}}^{(k,i)} \leq \tilde{B}_{k} = \sqrt{\sum_{j \in \pi(k)}\tilde{p}_{j,i}^{2}} \leq \tilde{p}_{\mathrm{max}}^{(k,i)}
	\end{equation}
	
	\noindent where $\tilde{p}_{\mathrm{max}}^{(k,i)}$ and $\tilde{p}_{\mathrm{min}}^{(k,i)}$ are defined similarly with $p_{\mathrm{max}}^{(k,i)}$ and $p_{\mathrm{min}}^{(k,i)}$, \ie, $\tilde{p}_{\mathrm{max}}^{(k,i)} = \underset{j \in \pi(i)}{\max}\tilde{p}_{j,i}$, $\tilde{p}_{\mathrm{min}}^{(k,i)} = \underset{j \in \pi(k)}{\min}\tilde{p}_{j,i}$. Then, we have,
	
	\begin{equation}
		\frac{\sup {\lVert \tilde{\nabla}_{b,k} \rVert}}{\sup {\lVert \tilde{\nabla}_{b,l} \rVert}} = \frac{\frac{n_{i}}{n_{k}} \cdot \tilde{B}_{k}\sqrt{n_{k}E_{H,k}}}{\frac{n_{i}}{n_{l}} \cdot \tilde{B}_{l}\sqrt{n_{l}E_{H,l}}} \leq \tilde{C}_{kl} \cdot \sqrt{\frac{E_{H,k}}{E_{H,l}}}
	\end{equation}
	
	\noindent where $\tilde{C}_{kl} = \tilde{p}_{\mathrm{max}}^{(k,i)}/\tilde{p}_{\mathrm{min}}^{(k,i)}$ is a constant.
		
\end{proof}

\subsection{Proof of \cref{prop:2}}

\begin{proof}

	\begin{equation}
		\begin{aligned}
			\lVert \nabla_{a}^{(i)} \rVert & = \lVert \sum_{j \in \pi(i)}(1 - p_{j,i}) \boldsymbol{h}_{j} \rVert \\
			& \leq \sum_{j \in \pi(i)} (1 - p_{j,i}) \lVert \boldsymbol{h}_{j} \rVert \\
			& \leq \sqrt{\sum_{j \in \pi(i)}{(1 - p_{j,i})}^{2}}\sqrt{\sum_{j \in \pi(i)}{\lVert \boldsymbol{h}_{j} \rVert}^{2}} \\
			& = A_{k} \sqrt{n_{i}\overline{{\lVert \boldsymbol{h}_{(i)} \rVert}^{2}}} \leq A_{k} \sqrt{n_{i}E_{H,i}},
		\end{aligned}
	\end{equation}
	
	\noindent where 
	
	\begin{equation}
		A_{k} = \sqrt{\sum_{j \in \pi(i)}{(1 - p_{j,i})}^{2}}.
	\end{equation} 
	
	Similar to the proof of \cref{prop:2},
	
	\begin{equation}
		\lVert \nabla_{b,k}^{(i)} \rVert = \lVert \underset{j \in \pi(k)}{\sum} p_{j,i} \boldsymbol{h}_{j} \rVert \leq B_{k} \sqrt{n_{k}\overline{{\lVert \boldsymbol{h}_{(k)} \rVert}^{2}}} \leq B_{k} \sqrt{n_{k}E_{H,k}},
	\end{equation}
	
	For $j \in \pi(i)$, the maximum of $p_{j,i}$ is less than $1$ and we denote this maximum as $p_{\mathrm{max}}^{(i)}$. For $j \in \pi(k), \; k \neq i$, there is a non-zero minimum for $p_{j,i}$ and we denote this minimum as $p_{\mathrm{min}}^{(k,i)}$. Then we have,
	
	\begin{equation}
		A_{k} \geq \sqrt{n_{i}} (1 - p_{\mathrm{max}}^{(i)}), \quad B_{k} \leq \sqrt{n_{k}}p_{\mathrm{min}}^{(k,i)}.
	\end{equation}
	
	\noindent Therefore, the proportion of the supremum of the attraction and $k$-th($k \neq i$) repulsion terms' norm obey:
	
	\begin{equation}
		\frac{\sup {\lVert \nabla_{a}^{(i)} \rVert}}{\sup {\lVert \nabla_{b,k}^{(i)} \rVert}} = \frac{A_{k} \sqrt{n_{i}\overline{{\lVert \boldsymbol{h}_{(i)} \rVert}^{2}}}}{B_{k} \sqrt{n_{k}\overline{{\lVert \boldsymbol{h}_{(k)} \rVert}^{2}}}} \leq \frac{n_{i}}{n_{k}} \cdot C_{k} \cdot \sqrt{\frac{E_{H,i}}{E_{H,k}}},
	\end{equation}
	
	\noindent where $C_{k} = (1 - p_{\mathrm{max}}^{(i)}) / p_{\mathrm{min}}^{(k,i)}$ is a constant.
	
	Similarly, for the ARB-Loss, 
	
	\begin{equation}
		\begin{aligned}
			\lVert \tilde{\nabla}_{a}^{(i)} \rVert & = \lVert \sum_{j \in \pi(i)}(1 - \tilde{p}_{j,i}) \boldsymbol{h}_{j} \rVert \\
			& \leq \sum_{j \in \pi(i)}(1 - \tilde{p}_{j,i}) \lVert \boldsymbol{h}_{j} \rVert \\
			& \leq \sqrt{\sum_{j \in \pi(i)}(1 - \tilde{p}_{j,i})^{2}} \sqrt{\sum_{j \in \pi(i)}{\lVert \boldsymbol{h}_{j} \rVert}^{2}} \\
			& = \tilde{A}_{k} \sqrt{n_{i}\overline{{\lVert \boldsymbol{h}_{(i)} \rVert}^{2}}} \leq \tilde{A}_{k} \sqrt{n_{i}E_{H,i}} \\
		\lVert \tilde{\nabla}_{b,k} \rVert & = \lVert \frac{n_{i}}{n_{k}}\sum_{j \in \pi(k)}\tilde{p}_{j,i} \boldsymbol{h}_{j} \rVert \\
			& \leq \frac{n_{i}}{n_{k}}\sum_{j \in \pi(k)}\tilde{p}_{j,i} \lVert \boldsymbol{h}_{j} \rVert \\
			& \leq \frac{n_{i}}{n_{k}}\sqrt{\sum_{j \in \pi(k)}\tilde{p}_{j,i}^{2}} \sqrt{\sum_{j \in \pi(k)}{\lVert \boldsymbol{h}_{j} \rVert}^{2}} \\
			& = \frac{n_{i}}{n_{k}}\cdot\tilde{B}_{k}\sqrt{n_{k}{\overline{\lVert \boldsymbol{h}_{(k)} \rVert}}^{2}} \leq \frac{n_{i}}{n_{k}}\cdot\tilde{B}_{k}\sqrt{n_{k}E_{H,k}}
		\end{aligned},
	\end{equation}
	
	\noindent where,
	
	\begin{equation}
		\begin{aligned}
			\tilde{A}_{k} & = \sqrt{\sum_{j \in \pi(i)}(1 - \tilde{p}_{j,i})^{2}} \geq \sqrt{n_{i}} (1 - \tilde{p}_{\mathrm{max}}^{(i)}) \\
			\tilde{B}_{k} & = \sqrt{\sum_{j \in \pi(k)}\tilde{p}_{j,i}^{2}} \leq \sqrt{n_{k}}\tilde{p}_{\mathrm{min}}^{(k,i)}
		\end{aligned}
	\end{equation}
	
	\noindent where $\tilde{p}_{\mathrm{max}}^{(i)}$ and $\tilde{p}_{\mathrm{min}}^{(k,i)}$ are defined similarly with $p_{\mathrm{max}}^{(i)}$ and $p_{\mathrm{min}}^{(k,i)}$, \ie, $\tilde{p}_{\mathrm{max}}^{(i)} = \underset{j \in \pi(i)}{\max}p_{j,i}$, $\tilde{p}_{\mathrm{min}}^{(k,i)} = \underset{j \in \pi(k)}{\max}p_{j,i}$. Therefore,
	
	\begin{equation}
		\frac{\sup {\lVert \tilde{\nabla}_{a}^{(i)} \rVert}}{\sup {\lVert \tilde{\nabla}_{b,k} \rVert}} = \frac{\tilde{A}_{k} \sqrt{n_{i}\overline{{\lVert \boldsymbol{h}_{(i)} \rVert}^{2}}}}{\frac{n_{i}}{n_{k}}\cdot\tilde{B}_{k}\sqrt{n_{k}{\lVert \boldsymbol{h}_{(k)} \rVert}^{2}}} \leq \tilde{C}_{k} \cdot \sqrt{\frac{E_{H,i}}{E_{H,k}}},
	\end{equation}
	
	\noindent where $\tilde{C}_{k} = (1 - \tilde{p}_{\mathrm{max}}^{(i)}) / \tilde{p}_{\mathrm{min}}^{(k,i)}$ is a constant.
	
\end{proof}

\section{Datasets}\label{appendix:dataset}

\subsection{Classification Datasets} 

We perform experiments on CIFAR-10/100 \cite{krizhevsky2009learning} and three large-scale image classification datasets: ImageNet \cite{deng2009imagenet}, iNaturalist2018 \cite{van2018inaturalist} and Places \cite{zhou2017places}. 

\textbf{CIFAR-10/100} are both composed of a training set of $50000$ images and a validation set of $10000$ images with $10$ and $100$ classes, respectively. The long-tail versions of CIFAR-10/100 are generated by the same rule as \cite{cao2019learning}, \ie, reducing the sample number per class according to an exponential curve controlled by an imbalance factor $\mathrm{IF} = \frac{n_{\mathrm{max}}}{n_{\mathrm{min}}}$, where $n_{\mathrm{max}}$ and $n_{\mathrm{min}}$ are the sample numbers of the most and the least frequent classes.

\textbf{ImageNet-LT} and \textbf{Places-LT} were first proposed by \cite{liu2019large}. ImageNet is composed of $1,281,167$ training images, $50,000$ validation images and $100,000$ test images and spans $1000$ classes. Places contains $10,624,928$ images from $434$ classes. \textbf{ImageNet-LT} has $115.8$K images from $1000$ classes whose cardinalities ranging from $5$ to $1,280$. \textbf{Places-LT} contains $184.5$K images from $365$ classes with class cardinality ranging from $5$ to $4,980$. 

\textbf{iNaturalist2018} is a dataset that originally suffers from an extreme class imbalance, which consists of $437.5$K images from $8,142$ classes. Besides, we also face the fine-grained classification in iNaturalist2018 dataset.

\subsection{Segmentation Datasets}

Essentially, the segmentation is a pixel-wise classification. And it usually suffers from extremely imbalanced distributions, e.g. the sky and ground usually occupy much more pixels than foreground objects on the image captured by the car-equipped camera. 

\textbf{CityScapes} \cite{cordts2016cityscapes} contains $5000$ annotated images of the street scenes of 50 different cities and $19$ semantic labels are used for evaluation. The training, validation, and test sets contain 2975, 500, and 1525 images, respectively.

\section{Implementation Details}\label{appendix:impl}

We use the PyTorch \cite{paszke2019pytorch} framework to perform all the experiments. For all experiments on the image classification task, we use an SGD optimizer with a momentum of $0.9$, and an initial learning rate of $0.1$. For experiments on CIFAR-10/100(-LT), we train the model on a single GPU, and the learning rate decays as its $0.1$ at the $160$-th and $180$-th epochs. For ImageNet-LT, Places-LT, and iNaturalist2018 datasets, we train the models on $8$ GPUs and use cosine learning rate schedule \cite{loshchilov10sgdr} gradually decaying from $0.2$ to $0$. For all experiments, we use mixup \cite{zhang2018mixup} data augmentation, and the batch size is $128$ per GPU. For experiments on the image segmentation task, we use an SGD optimizer with an initial learning rate of $0.01$, a momentum of $0.9$, and a weight decay of $0.0005$. And we update the learning rate by the poly policy with a power of $0.9$.

\section{Detailed Results of Image Segmentation Task}

\begin{table}[h]
\resizebox{\textwidth}{!}{
\begin{tabular}{c|c|c|ccccccccccccccccccc|c}
\hline
Backbone &
  Loss &
  Metric &
  road &
  sidewalk &
  building &
  wall &
  fence &
  pole &
  traff light &
  traff sign &
  vegetation &
  terrain &
  sky &
  person &
  rider &
  car &
  truck &
  bus &
  train &
  motor &
  bicycle &
  mean \\ \hline
\multirow{4}{*}{U-Net} &
  \multirow{2}{*}{\begin{tabular}[c]{@{}c@{}}ARB-\\ Loss\end{tabular}} &
  IoU &
  97.01 &
  78.92 &
  92.83 &
  39.73 &
  50.86 &
  61.05 &
  70.68 &
  75.34 &
  90.98 &
  55.96 &
  92.83 &
  78.12 &
  55.11 &
  95.46 &
  38.71 &
  59.17 &
  33.86 &
  49.98 &
  73.33 &
  67.89 \\
 &
   &
  Acc &
  97.86 &
  88.57 &
  93.37 &
  48.4 &
  60.18 &
  82.74 &
  94.05 &
  91.16 &
  94.41 &
  72.41 &
  98.96 &
  93.6 &
  65.77 &
  96.75 &
  48.21 &
  67.5 &
  60.94 &
  69.24 &
  89.93 &
  79.69 \\ \cline{2-3}
 &
  \multirow{2}{*}{\begin{tabular}[c]{@{}c@{}}Cross-\\ Entropy\end{tabular}} &
  IoU &
  96.82 &
  77.25 &
  90.05 &
  31.86 &
  45.77 &
  63.33 &
  73.74 &
  78.43 &
  91.89 &
  57.94 &
  94.38 &
  79.51 &
  48.82 &
  92.97 &
  33.63 &
  62.8 &
  34.65 &
  53.16 &
  76.63 &
  67.56 \\
 &
   &
  Acc &
  98.61 &
  86.47 &
  96.33 &
  35.54 &
  52.81 &
  73.4 &
  83.91 &
  85.31 &
  96.35 &
  68.11 &
  97.97 &
  92.0 &
  55.94 &
  97.44 &
  36.51 &
  70.04 &
  40.91 &
  61.39 &
  87.48 &
  74.55 \\ \hline
\multirow{4}{*}{deeplabv3+} &
  \multirow{2}{*}{\begin{tabular}[c]{@{}c@{}}ARB-\\ Loss\end{tabular}} &
  IoU &
  97.99 &
  89.97 &
  91.93 &
  58.15 &
  64.64 &
  67.51 &
  73.46 &
  78.18 &
  91.91 &
  63.1 &
  93.16 &
  82.36 &
  72.08 &
  96.62 &
  75.61 &
  86.64 &
  83.97 &
  64.86 &
  78.24 &
  79.49 \\
 &
   &
  Acc &
  98.49 &
  93.11 &
  94.62 &
  62.26 &
  72.19 &
  87.95 &
  95.64 &
  94.71 &
  94.89 &
  80.4 &
  99.3 &
  95.38 &
  80.06 &
  97.87 &
  85.26 &
  92.52 &
  90.28 &
  82.82 &
  92.33 &
  88.95 \\ \cline{2-2}
 &
  \multirow{2}{*}{\begin{tabular}[c]{@{}c@{}}Cross-\\ Entropy\end{tabular}} &
  IoU &
  98.31 &
  86.12 &
  93.1 &
  54.84 &
  61.31 &
  69.62 &
  75.11 &
  81.82 &
  92.9 &
  65.51 &
  95.2 &
  84.08 &
  63.63 &
  95.87 &
  77.36 &
  88.21 &
  75.41 &
  68.68 &
  80.17 &
  79.33 \\
 &
   &
  Acc &
  99.08 &
  93.07 &
  96.82 &
  59.6 &
  69.51 &
  81.66 &
  86.41 &
  89.41 &
  96.83 &
  74.68 &
  98.37 &
  92.93 &
  74.01 &
  98.12 &
  84.8 &
  96.04 &
  78.08 &
  79.29 &
  90.54 &
  86.28 \\ \hline
\end{tabular}
}
\label{tab:detail_seg}
\caption{The detailed results on CityScapes \cite{cordts2016cityscapes} dataset based on tow backbone networks: U-Net \cite{ronneberger2015u} and Deeplabv3+ \cite{chen2018encoder}.}
\end{table}

From the per-class results in the above table, we can see that the classes whose IoU gets improved while the pixel accuracy is still a little lower are almost the minor classes. The reason might be the differences in the definition of these two metrics. For a certain class $i$, IoU is defined as:

\begin{equation}
	\mathrm{IoU}_{i} = \frac{\mathrm{TP}_{i}}{\mathrm{TP}_{i} + \mathrm{FP}_{i} + \mathrm{FN}_{i}}
\end{equation}

And pixel accuracy is defined as:

\begin{equation}
	\mathrm{Acc}_{i} = \frac{\mathrm{TP}_{i}}{\mathrm{TP}_{i} + \mathrm{FN}_{i}}
\end{equation}

If the $\mathrm{FP}_{i}$ item is relatively too large for the cardinality of a minor class, \ie, $\mathrm{FP}_{i}$ is too large compared to $\mathrm{TP}_{i} + \mathrm{FN}_{i}$, the IoU value will drop a lot. Meanwhile, if these $\mathrm{FP}_{i}$ pixels belongs to some major classes, it will have little effect on the pixel accuracy of these major classes. So, it will cause that although pixel accuracy improves a lot, the improvements in the metric of IoU are slight.

\end{document}